\documentclass[11pt,a4paper]{article}
\usepackage[hyperref]{emnlp2018}
\usepackage{times}
\usepackage{latexsym}

\usepackage{url}
\usepackage{graphicx}
\graphicspath{ {images/} }

\aclfinalcopy 

\title{Detecting Satire in the News with Machine Learning}

\author{Andreas St\"ockl \\
{\tt andreas.stoeckl@fh-hagenberg.at} \\
       School of Informatics, Communications and Media\\
       University of Applied Sciences Upper Austria\\
       Softwarepark 11, 4232 Hagenberg, Austria  \\
}

\date{}

\begin{document}
\maketitle

\begin{abstract}

We built models with Logistic Regression and linear Support Vector Machines on  a large dataset consisting of regular news articles and news from satirical websites,
and showed that such linear classifiers on a corpus with about 60,000 articles can perform with a precision of 98.7\%  and a recall of 95.2\% on a random test set 
of the news.  

On the other hand, when testing the classifier on ``publication sources'' which are completely unknown during training, only an accuracy of 88.2\% and an F1-score of 76.3\%
are achieved.

As another result, we showed that the same algorithm can distinguish between news written by the news agency itself and paid articles from customers. Here the results
had an accuracy of 99\%.
\end{abstract}

\section{Introduction}

\cite{banko2001scaling} determined that, with some learning curve experiments for typical natural language classification tasks, the performance of
learners can benefit significantly from much larger training sets. 
They propose to direct efforts towards increasing the size of annotated training collections, while reducing the effort on comparing different
learning techniques and features trained only on small training corpora. 

\cite{halevy2009unreasonable} proposed that the biggest successes in natural-language-related machine learning have been
in domains in which a large training set is available.
For the feature selection task, we should trust that human language has already evolved words for the most important concepts.
Thus we can proceed by tying together the words that are already there, rather than by inventing new concepts. 

In this paper, we experiment with methods for training a machine learning system to detect satirical or fake news by analyzing
the text content of the article. 
The amount of online published news articles has been rapidly growing over the last decades and is still steadily increasing.
We assembled a new text corpus for training and testing from German news articles published over the last decade. 
The source for the non-fake articles are documents published by news agencies and newspapers on the web.
As training data for the fake news, we use documents published by satirical news websites, and so they are declared as fake and can be
used for supervised learning methods (see Section \ref{Dataset}).
The articles come from different domains such as politics, business, technology and others.

Sometimes it seems even difficult for humans to recognize the satirical or fake character of news articles.
We endeavor to test which level confidence an algorithm can achieve by analyzing the pure text content without any information like links to other articles.
For humans, to determine the trustworthiness of an article, it is very important to know who published the content. Humans learn to trust some news sources and 
not others. We will test how machine learning performs if the systems knows the sources in the sense that it was trained on texts published there.
To see if the system can detect satirical content without knowing any articles from the publisher, we also test on a set of articles from publishers we did not use in any way for training.

\section{Related work}

There are numerous linguistic approaches for fake news detection which use simple methods of representing texts, such as the “bag of words” or “n-grams” or analyzing deeper language structures.

And there are network approaches which use the network structure of linked data or social network behavior. A good overview can be found in \cite{PRA2:PRA2145052010082}.
The goal in the linguistic approaches is to find deception cues like frequencies and patterns of pronoun, conjunction, and negative emotion word usage. \cite{feng2013detecting}
worked with n-grams features and deep syntax features to differentiate genuine and fake product reviews.

An example of a network approach can be found in \cite{DBLP:journals/corr/TacchiniBVMA17} in which Facebook posts are classified as hoaxes or non-hoaxes on the basis of the users
who “liked” them. They used logistic regression and other techniques on a dataset consisting of 15,500 Facebook posts and 909,236 users, and they obtained
classification accuracies exceeding 99\%.

\cite{DBLP:journals/corr/Wang17j} presented a dataset for fake news detection consisting of 12.8K manually labeled short statements in various contexts from PolitiFact.com.
They implemented a hybrid approach with a convolutional neural network to integrate meta-data with text.
\cite{amador2017characterizing} were characterizing political fake news on Twitter through the analysis of their meta-data. 

\cite{rubin2016fake} analyzed a very similar setup as we do, by looking at news articles from the newspapers {\em The Toronto Star} and {\em The New York Times}
and the satirical news sites {\em The Onion} and {\em The Beaverton}. They did so by looking only at a small set of articles (480 documents) and searching for handcrafted features which can be used for regression methods or support vector machines. They found a precision of 0.78 and a recall of 0.87 and therefore an F1-score of 0.82 for a baseline implementation using only 
tf-idf topic vectors. They tried out different additional features in their setup suggested from past literature, and found that they could improve performance slightly to an F1-score of 0.87.
The features tested were combinations of things like the presence of new named entities at the end of the article, the punctuation, the percentage of linguistic categories in the text or
or topic and sentiment classification.

The training and evaluation were done with support vector machines (SVMs) using 75\% of the articles as training data with 10-fold cross validation and 25\% as test set.
The test set was a random sample from documents from a mixture of all the publishers which were used for training and testing.

A very similar study to ours was made by \cite{ahmed2017detection} with a dataset of news articles from reuters.com and a fake news dataset from PolitiFact.com collected by \cite{DBLP:journals/corr/Wang17j}. They tested different supervised classification techniques and found best results for linear SVMs with an accuracy of 92\%.
\cite{8100379} tried a simple naive Bayes classifier for fake news detection a data set of Facebook news posts and  accuracy of approximately 74\%.

\cite{pisarevskaya2017deception} used SVMs (linear/ rbf kernel) and Random Forest to classify a corpus of 174  Russian news reports (truthful and fake ones) 
They used frequencies of POS tags and frequencies of rhetorical relations types in texts and got F1-measure of 65\% with SVMs and
 F1-measure of 54\% with Random Forest Classifier.

\section{Dataset and Data Collection} \label{Dataset}

\begin{table}
\small
\centering
\begin{tabular}{|l|l|r|}
\hline \bf URL & \bf Date &  \bf No. Articles \\ \hline
de.reuters.com & 2015 to 2018 & 32,984 \\
www.pressetext.com &  2008 to 2018 & 22,896 \\
www.kleinezeitung.at & 2018 & 3,361  \\
\hline
\end{tabular}
\caption{News agencies and newspapers}
\label{table:1}
\end{table}

We collected datasets of news articles in German language from news agencies and newspapers via their websites (Table \ref{table:1}), and from satirical news sites (Table \ref{table:2}) for training and testing of the model.

\begin{table}
\small
\centering
\begin{tabular}{|l|l|r|}
\hline \bf URL & \bf Date & \bf No. Articles \\ \hline
www.der-postillon.com &  2008 to 2018 & 2,040 \\
www.der-zeitspiegel.de &  2010 to 2018 & 135 \\
www.eine-zeitung.net & 2010 to 2018 & 1,322  \\
www.dietagespresse.at & 2010 to 2018 & 1,130  \\
\hline
\end{tabular}
\caption{Satirical news sites}
\label{table:2}
\end{table}

After downloading some data cleaning, routines were performed on the data, deleting information such as marks of the publisher, author names and some footers or text-boxes - all pieces of text on which the machine learning could learn to detect the source without analyzing the content of the article. And all publishing dates were converted to a common format.

\cite{PRA2:PRA2145052010083} stated requirements for a fake news detection corpus based on reviewed practices and their own research.
The ``availability of both truthful and deceptive instances'' and ``verifiability of ground truth'' was done by the choice of satirical news websites and normal news agencies and newspapers.
For ``digital textual format accessibility'' we downloaded the documents to a local SQL-Database. 

``Homogeneity in lengths'' was achieved in a preprocessing step by deleting articles with less than 500 characters or more than 10,000 characters in the body text. 
For ``Homogeneity in writing matter and time'' we stored the category in which the article was published and the date of publishing.

For each article we stored:
\begin{itemize} 
\item The {\tt URL} at which the article was published
\item The {\tt  title}
\item The {\tt body text}
\item A {\tt category} that was given to the article by the publisher
\item The {\tt date} of publishing
\item The {\tt publisher} and therefore if satirical or not
\end{itemize}
In total we collected 63,868 articles in a local database, which then was used to select the data for training and testing.

\section{Training and model selection}  \label{Model}

The text preprocessing pipeline was  scripted in Python and used the {\em scikit-learn} open source machine learning package \cite{pedregosa:hal-00650905, geron2017hands, raschka2017python}.
First the raw text of the articles and the title was imported to a Pandas\footnote{\label{foot:pd} https://pandas.pydata.org/} data frame \cite{mckinney2012python} and a pipeline for count-vectorization, tf*idf weighting, stop-word removal, uni-gram and bi-word tokenization was built. For classification we used Logistic Regression and Support Vector Machines.
 
\begin{figure}
	\centering
	\includegraphics{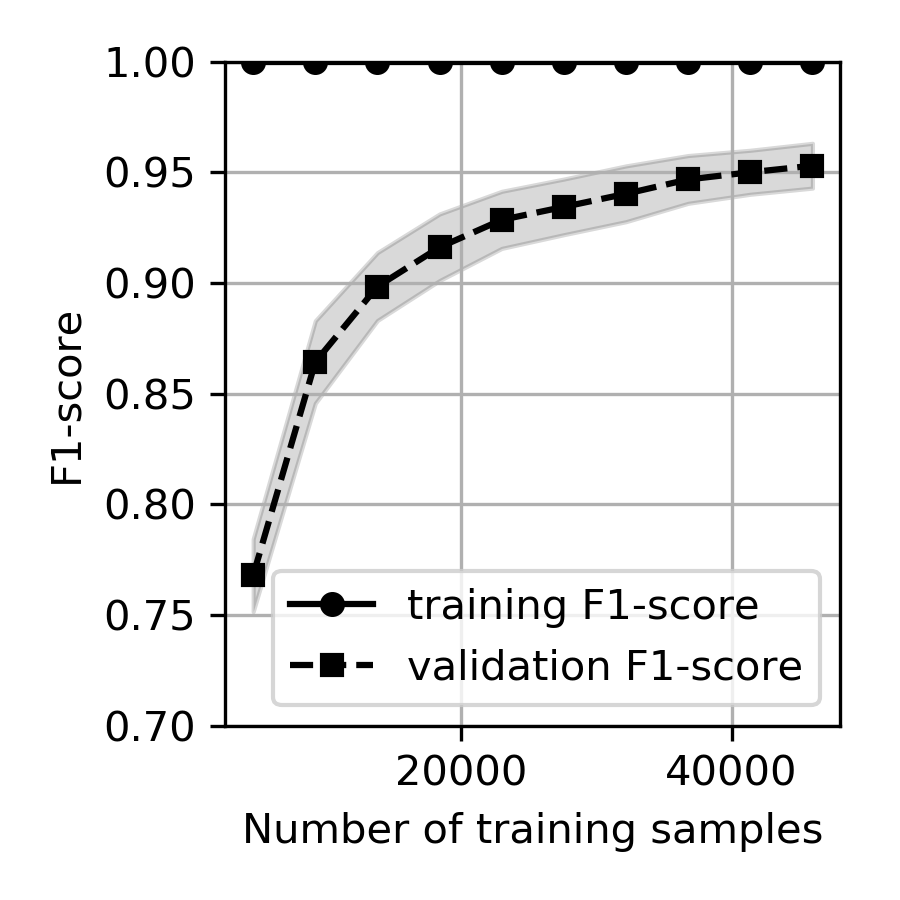}
	\caption{Learning curve of a 80\% training set with Logistic Regerssion and C=1 }
	\label{img:learningcurve}
\end{figure}

First we split the whole dataset randomly into 80\% training data and the rest for testing.
We then started with standard Logistic Regeression and regularisation parameter $C=1$ and plotted the learning curves (Figure \ref{img:learningcurve}) with 10-fold cross validation.
Since the number of fake examples in the dataset is much lower than the number of regular news, we use the F1-score as target measure.
After a suitable number of training-examples the F1-score reaches a  performance of approximately 85\%. 

Next we tried to optimize the regularisation parameter $C$ to improve the performance. A grid-search shows that high values of C and therefore weak regularisation produces better results.

In the vectorization step, best results are achieved with uni-grams and bi-grams converted to lowercase. We use no classic stop-word filtering list, but terms that are used in more than 80\% of the documents are filtered, and only terms that occur 20 or more times are used for the dictionary.

\begin{figure}
	\centering
	\includegraphics{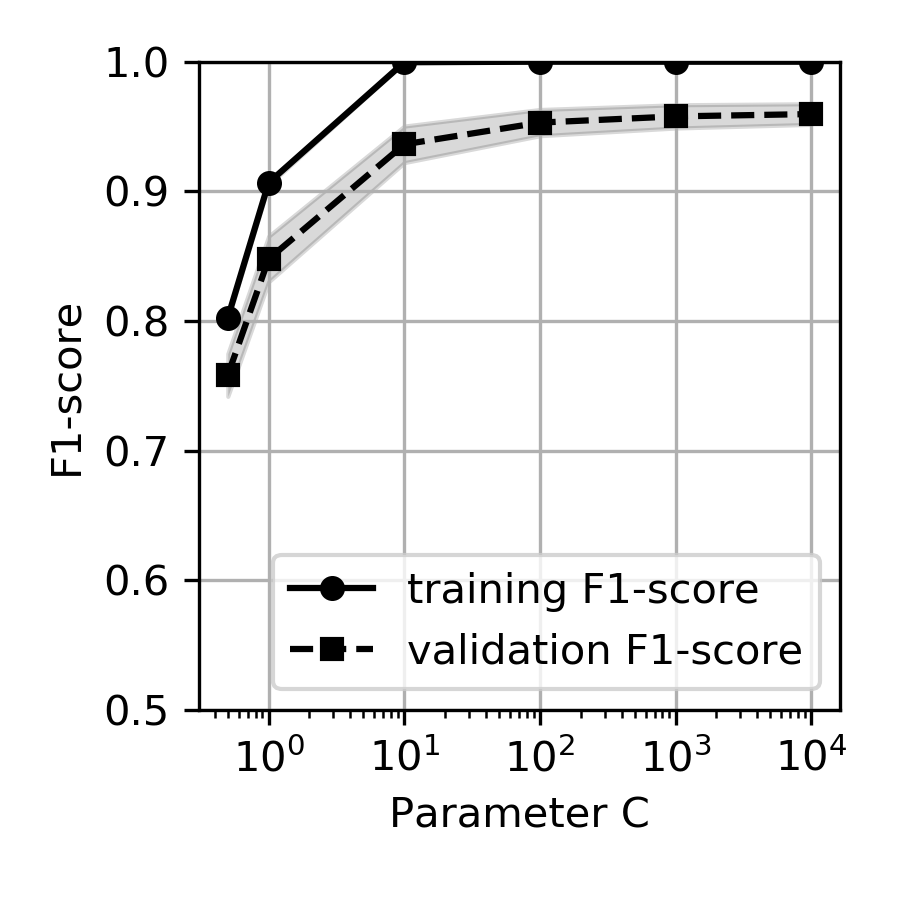}
	\caption{Validation curve for parameter C of Logistic Regerssion }
	\label{img:validationcurve}
\end{figure}

For Logistic Regression, the regularisation parameter C was set to 1,000 (Figure \ref{img:validationcurve}) and for Support Vector Machines (SVM) C=100. SVMs with linear kernel produces slightly better results with more computorial cost than Logistic Regression. The nonlinear kernels ``rbf'' and ``sigmoid'' did not work in our setup due to overfitting. So we will use linear SVMs with the preprocessing and parameters described above for the results.

The vocabulary has a length of 125,623 terms consisting of uni-grams and bi-grams.

\section{Results}

\begin{figure}
	\centering
	\includegraphics{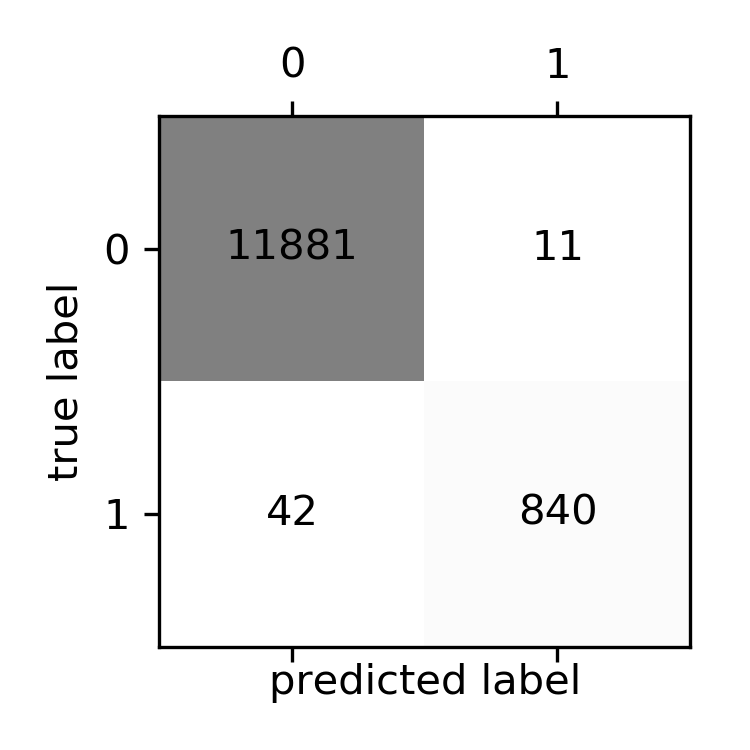}
	\caption{Confusion matrix of SVM with C=100 for a random test set}
	\label{img:confusionmatrix}
\end{figure}

With the model derived in section \ref{Model}, we tested the classifier on the test set of randomly selected 20\% of the corpus.
On the set of 11,892 regular news and 882 satirical news, it classifies 11 regular news incorrectly and did not detect 42 of the fake news articles
(Figure \ref{img:confusionmatrix} shows the confusion matrix).

This gives a test accuracy of 0.996, a precision of 0.987, and a recall of 0.952 and so an F1-score of 0.969.
This is an improvement against the  F1-score of 0.87 of \cite{rubin2016fake} which was performed with 480 English news and satirical news articles and
similar to the results from \cite{DBLP:journals/corr/Wang17j} with 92\% accuracy but a more balanced dataset then ours.
This shows the possibilities of a linear model with a lot of data without any handcrafted features.

But are we sure that the classifier detects the satire in the news? Or does it learn to identify the publisher of the article?
For the training of a classifier for the ``publication source'' we use again a SVM with the same hyperparameter as in section \ref{Model}.
The six different classes of publishers were predicted fairly precisely on a 20\% random test set with an accuracy of 98.4\%.
So "knowing" articles from all publishers gives the algorithm clues for detecting satire by learning the features for detecting the publisher.

\begin{figure}
	\centering
	\includegraphics{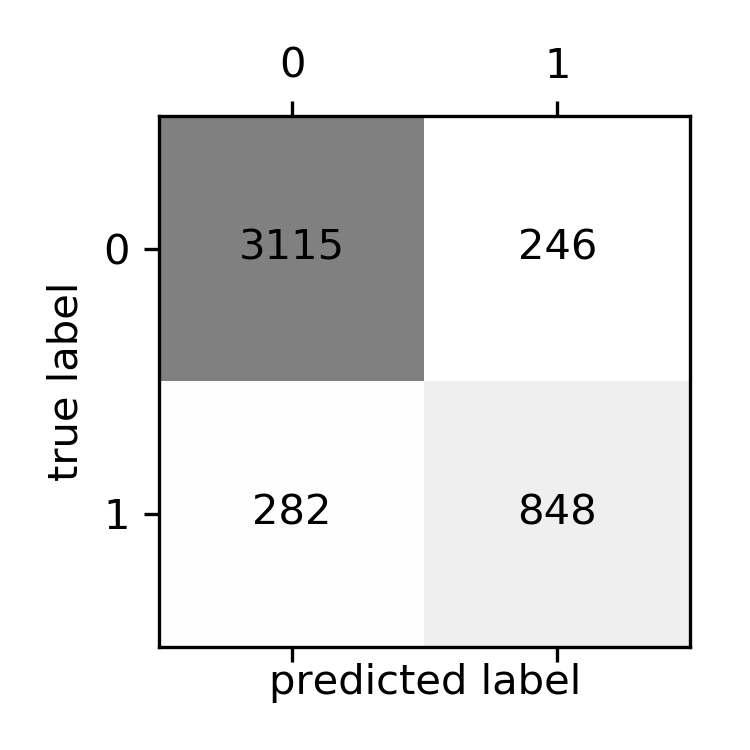}
	\caption{Confusion matrix of SVM with C=100 for "unknown" publishers}
	\label{img:confusionmatrix2}
\end{figure}

Thus, to check if the classifier can detect the satire without knowing examples from a publisher, we tried a different setup.
We excluded two sources of news, one regular news site (``Kleine Zeitung'') and one satirical (``Die Tagespresse'') completely from the training dataset, and then used these articles as a test set.
This provides a set of 3,361 regular and 1,130 satire test articles. The setup should be much harder for the classifier. And of course the results of a SVM classification with the same hyperparameters as configured in section \ref{Model} produces different results.
The satire was detected with an accuracy of 88.2\%, a precision of 77.5\% , recall of 75\% and therefore F1-score of 76.3\%. (Figure \ref{img:confusionmatrix2} shows the confusion matrix)

Another interesting question arose while examining the data of the publisher ``Pressetext Austria''. There are different types of articles. Some are written by the news agency and some
are paid by customers as PR text.
This is marked in the online publication, so we can extract this information and use it for another classification.
Is this type of news a sort of ``lie''? Can the algorithm detect which article was paid?

There are 14,191 number of regular news from this publisher and 7,373 paid articles.
On a 20\% random test set, the two types of articles are classified with an accuracy of 99\%.

\section{Future Work}

We think the presented methods can be used with other languages  and we expect similar results as with the German corpus.
Using non-linear methods for classification could improve the performance, but would need a much larger training set as we had.
Our experiments with non-linear kernels in SVMs demonstrated bad results due to overfitting.

It would be interesting to use a classifier trained on satire fake news on a test set of fake news from non-satirical sources. 
A set of such news would be needed for this, and as indicated from the results on the untrained news sources, we expect satisfactory results only if enough articles from all sources are in the training set.

\bibliography{satire}
\bibliographystyle{acl_natbib_nourl}

\appendix

\end{document}